\documentclass[sn-chicago]{sn-jnl}% Chicago-based Humanities Reference Style
\usepackage{graphicx}
\usepackage{epsfig} % for postscript graphics files
\usepackage{subfigure}
\usepackage{amsmath} 
\usepackage{enumerate}
\usepackage{ulem}
\usepackage{natbib}

\jyear{2022}%

\theoremstyle{thmstyleone}%
\theoremstyle{thmstyletwo}%

\theoremstyle{thmstylethree}%

\begin{document}

\title[Siamese Neural Networks for robot localization in indoor environments]{An experimental evaluation of Siamese Neural Networks for robot localization using omnidirectional imaging in indoor environments}


\author*[1]{\fnm{Juan Jos\'e} \sur{Cabrera}}\email{juan.cabreram@umh.es}
\author[1]{\fnm{Vicente} \sur{Rom\'an}}%\email{sergio.cebollada@umh.es}
\author[1]{\fnm{Arturo} \sur{Gil}}%\email{m.flores@umh.es}
\author[1,2]{\fnm{Oscar} \sur{Reinoso}}%\email{o.reinoso@umh.es}
\author[1]{\fnm{Luis} \sur{Pay\'a}}%\email{lpaya@umh.es}

\affil[1]{\orgdiv{Institute for Engineering Research (I3E)}, \orgname{Miguel Hern\'andez University}, \orgaddress{\city{Elche}, \country{Spain}}}

\affil[2]{\orgdiv{Valencian Graduate School and Research Network for Artificial Intelligence (valgrAI)}, \orgaddress{\country{Spain}}}


\abstract{The objective of this paper is to address the localization problem using omnidirectional images captured by a catadioptric vision system mounted on the robot. For this purpose, we explore the potential of Siamese Neural Networks for modeling indoor environments using panoramic images as the unique source of information. Siamese Neural Networks are characterized by their ability to generate a similarity function between two input data, in this case, between two panoramic images. In this study, Siamese Neural Networks composed of two Convolutional Neural Networks (CNNs) are used. The output of each CNN is a descriptor which is used to characterize each image. The dissimilarity of the images is computed by measuring the distance between these descriptors. This fact makes Siamese Neural Networks particularly suitable to perform image retrieval tasks. First, we evaluate an initial task strongly related to localization that consists in detecting whether two images have been captured in the same or in different rooms. Next, we assess Siamese Neural Networks in the context of a global localization problem. The results outperform previous techniques for solving the localization task using the COLD-Freiburg dataset, in a variety of lighting conditions, specially when using images captured in cloudy and night conditions.}




\maketitle

\section{Introduction}

During the past few years, vision sensors have been used extensively in the field of map building and localization with mobile robots \citep{hu2020multi,zhong2018detect}. In particular, the ability to localize in the map is of paramount importance in order to develop autonomous robots that can navigate in real operation conditions  \citep{reinoso2020specialAS}. The interest in using vision sensors to capture information from the environment is still high. Cameras can capture a big amount of information from the environment with a relatively low cost and they can be used in both, indoor and outdoor areas. Additionally, the images permit carrying out other highly specialized tasks such as object recognition and people detection. 
\vspace{0.5cm}

Among the available configurations to capture visual information, the use of omnidirectional vision sensors in mobile robotics has become common. Omnidirectional cameras obtain images that cover a field of view of 360 \textit{deg.} around the robot \citep{junior2016calibration}. As a result, they are commonly used to address navigation tasks \citep{rituerto2010visual}.

% \textcolor{red}{The information captured with these systems can be projected onto different surfaces,  yielding different mathematical approaches depending on the type of task to solve \citep{coors2018spherenet}. }

\vspace{0.5cm}
The large amount of information provided by cameras requires robust techniques to extract and describe the relevant visual information.  Different paradigms have been considered to extract this relevant information. A first group of techniques concentrate on detecting, describing and tracking some landmarks or local features along the scenes \citep{garcia2015vision,luthardt2018llama,shamsfakhr2019indoor,cao2020study,lin2020orb}. Different local features have been used in mapping and localization tasks, including SIFT, SURF \citep{gil2009mva}  and ORB descriptors \citep{Rublee2011}. A global description of each image can then be obtained, for example, by means of the Bag of Words model \citep{murartal2015}. A second group of techniques work with each scene as a whole, and build a unique descriptor per image that contains information on its global appearance \citep{berenguer2015position,korrapati2017multi,khaliq2019holistic,liu2012visual}. Finally, hardware developments have led many authors to use Artificial Intelligence (AI) techniques to extract relevant information from images. Specifically, Convolutional Neural Networks (CNNs) have been proposed to address different computer vision and robotics tasks. For example, \cite{xu2019robust} and \cite{leyva2019tb} proposed global appearance descriptors based on a CNN to obtain the most probable pose of the robot. 

\vspace{0.5cm}

In general terms, holistic description methods lead to maps in which a set of robot poses and their associated descriptors are stored. In this way, each pose of the robot is represented by a holistic descriptor and this representation leads to straightforward localization algorithms, based on the pairwise comparison between descriptors.

% \textcolor{red}{ The localization can be addressed using two different strategies: hierarchical or global localization. On the one hand, the hierarchical localization is performed in two steps: (1) identifying the room where the test image is captured, and (2) retrieving the most similar pose among the images in that room. On the other hand, the global localization solves the localization problem in a single step comparing the test image directly with all the images in the visual model.}
\vspace{0.5cm}

In this manuscript we assess the usage of Siamese Neural Networks in the context of image description and robot localization. Siamese Neural Networks permit evaluating two images at the same time in such a way that they provide a similarity measurement at the output. Therefore, they have the potential to address visual recognition of places and estimate the position of a mobile robot. In the present paper, we evaluate this potential. The main contributions of this paper can be summarized as follows.

\vspace{0.5cm}

\begin{enumerate}
\item We explore the capability of Siamese Neural Networks for modeling indoor environments, using panoramic images as the unique source of information.
\item We train and evaluate Siamese Neural Networks with the purpose of detecting whether two images have been captured in the same or in different rooms.
\item We train Siamese Neural Networks capable of estimating robot position as a global image retrieval problem.
\item We conduct an exhaustive study on the influence of the Siamese Neural Networks' architecture and the most relevant parameters. Moreover, we analyze the robustness against some common visual phenomena that may occur in real operation conditions, such as changes of the lighting conditions or image blur.
\end{enumerate}

\section{State of the art}
\label{sec:RelatedWork}

As stated before, Siamese Neural Networks are able to generate a similarity function from pairs of input data. They can be regarded as a superstructure that includes two Neural Networks. These architectures accept two different inputs and offer a single output. The underlying networks share the same weights and different functions can be used to conform a single output. They were first proposed in 1993 in order to distinguish correct signatures from forgeries \citep{bromley1993}. Since then, these architectures have been proposed in different areas of knowledge. For example, \cite{thiolliere2015} proposed a Siamese Neural Network for audio and speech signal processing, \cite{zheng2019} used this architecture for the comparison of DNA sequences or \cite{jeon2019} used it for drug discovery purposes. Furthermore, \cite{parajuli2017} developed a Siamese Neural Network to track cardiac motion and \cite{sandouk2017} proposed a Siamese architecture in order to recognize music tags. 

\vspace{0.5cm}

 During the past few years, AI in general and CNNs in particular have been used in the field of mobile robotics for a variety of purposes. For instance, for \textit{mapping} \citep{sinha2018convolutional,moolan2019improving,brahmbhatt2018geometry}, \textit{localization} \citep{weinzaepfel2019visual,li2017indoor,cattaneo2019cmrnet}, \textit{navigation} \citep{zhao2018cnn,ma2019using} and \textit{simultaneous localization and mapping} \citep{lu2019deep,liu2019efficient}. 
A complete state-of-the-art review on mobile robotics tasks based on the use of AI can be found in \citep{cebollada2020state}. Other applications of AI in the context of mobile robotics include:  \textit{self-driving navigation} \citep{sharma2017securing,polvara2018obstacle,organisciak2020unifying}, \textit{face detection and recognition} \citep{wang2017face,jiang2017face,hu2021facial}, \textit{object recognition and categorization} \citep{nozawa20213d,zaki2019viewpoint,feng2020resolving} and \textit{mapping and localization} \citep{tanzmeister2014grid,holliday2018scale,ruan2019mobile}.

\vspace{0.5cm}

Convolutional Neural Networks (CNNs) are the most popular techniques among  AI tools. Currently, they are used in many mapping and localization tasks due to their successful performance in many practical applications. They are designed to receive images as input and their structures are specially created to obtain descriptors that synthesize the information in them \citep{chollet2018deep}. Therefore, they can be used to describe the global appearance of an image. In this sense, \cite{cebollada2019hierarchical} proposed holistic descriptors obtained with a CNN to perform localization within topological models, studying their strength against illumination variations. Also, \cite{xu2019robust} and \cite{leyva2019tb} proposed these techniques to obtain the most probable robot position. Additionally, \cite{ballesta2021cnn} studied localization tasks using CNNs and regression layers as global appearance descriptors. Some well known architectures have been used as basic structures to develop new modified networks for robotic navigation purposes. AlexNet \citep{krizhevsky2012imagenet},  VGG16 \citep{simonyan2014VVG}, GoogleNet \citep{szegedy2015going} or NetVLAD \citep{arandjelovic2016netvlad} are some of them.

\vspace{0.5cm}
The Convolutional Neural Networks presented above can be used to form a Siamese Neural Network. In the field of robotics, they have been rarely used and some studies that proposed this structure in this field are mentioned below. For example, \cite{utkin2017} use a Siamese Neural Network to support the security control of a robot by detecting anomalies in its behaviour and \cite{zeng2018} present a robotic pick-and-place system capable of identifying and grasping both known and novel objects in cluttered environments using a Siamese Neural Network. Moreover, \cite{li2019siamvgg} use the VGG16 network to conform a Siamese structure for object detection and tracking. Additionally, \cite{zhang2019deeper} presented a study in which Siamese Networks are followed by Fully Connected layers or Region Proposal Network structures in the context of real-time visual tracking. 

\vspace{0.5cm}
Regarding robot localization tasks, Leyva-Vallina \textit{et al.} have proposed the use of Siamese Neural Networks to address the place recognition problem in garden environments \citep{leyva2019placeSiamese,leyva2021generalized}. Moreover, this architecture has been proposed for localization using LiDAR scans \citep{yin2018locnet,chen2022}.

% Regarding the use of Siamese Neural Networks in mobile robotics, a detailed study is needed in order to know how they work in modelling and localization tasks, and which are the most appropriate configurations to address these problems. Leyva-Vallina \textit{et al.} has proposed this model to address the place recognition problem in garden environments \citep{leyva2019placeSiamese,leyva2021generalized}. Moreover, this kind of architecture has been proposed for localization departing from LiDAR scans \citep{yin2018locnet,chen2022}.

\vspace{0.5cm}

In the present paper, we address the localization of a mobile robot using panoramic images in such a way that we study in detail different architectures and training configurations of Siamese Neural Networks. For this purpose, we propose as an initial approach to train and test the capability of the network to distinguish between images captured in the same and different rooms. In addition, in this study we also tackle the global localization problem using Siamese Neural Networks.

\section{Siamese Neural Networks}
\label{sec:siamese_neuronal_networks}

\begin{figure}[]
\vspace{-0.2cm}
\centering
\includegraphics[width=\linewidth]{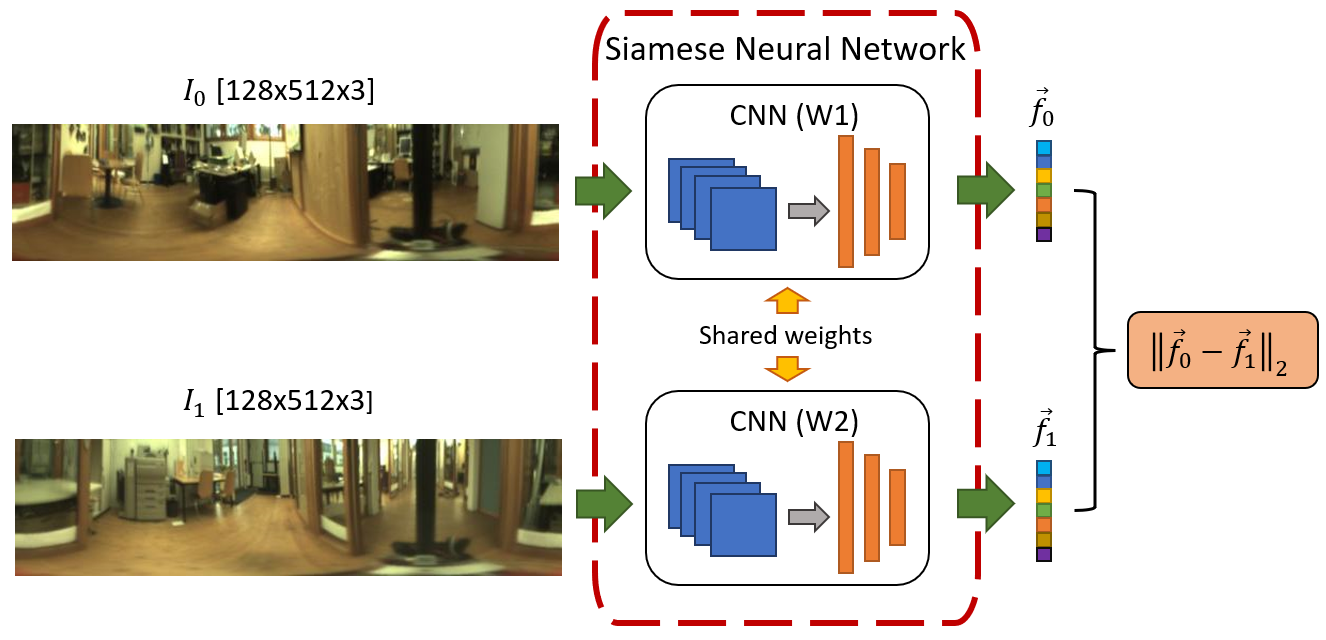}

%   {\epsfig{file = SiameseArquitecture2.PNG, width = 14cm}}
  \caption{Representation of the architecture of a general Siamese Neural Network.}
  \label{fig:siamese_arq}
  \vspace{-0.1cm}
\end{figure}

Siamese Neural Networks can be described as a superstructure that includes, at least, two different Neural Networks beneath. Weights are shared between the networks and a single output is generated by combining the outputs of both networks.  Figure \ref{fig:siamese_arq} shows a general representation of a Siamese Neural Network architecture. In the present work, we use Convolutional Neural Networks to conform the two branches of the Siamese Neural Network. The output of each CNN is a descriptor which is used to characterize each input image. The dissimilarity of the input images is computed by measuring the distance between these descriptors. In this way, Siamese Neural Networks can be trained to generate similar descriptors when the training images belong to the same category.  This fact makes Siamese Neural Networks particularly suitable to perform image retrieval tasks. Additionally, it is worth noting that the outputs, training, and performance of the network depend directly on:
\begin{itemize}
    \item The architectures used in subnetworks W1 and W2 to extract the main features of the images.
    \item The conversion of the feature maps from the convolutional layers to a descriptor vector.
    \item The dimension of the output descriptors that embed the pair of input images.
    \item The training carried out with the available images. In particular, the labelling and the ratio of images of each category.
    
\end{itemize}

In this manuscript, we analyze the influence of these items on the visual localization of the robot.

\unskip

% \begin{figure}[<placement-specifier>]
% \centering
% \includegraphics{<eps-file>}
% \caption{<figure-caption>}\label{<figure-label>}
% \end{figure}

\section{Visual Localization}
\label{sec:Visual_Localization}

This section is focused on explaining the mobile robot localization using visual information. In this manuscript, we assume that a visual map of the environment is initially available. To obtain this map, the robot has moved throughout the area capturing omnidirectional images along the trajectory. Firstly, the images are transformed to a panoramic format (with size 128x512 in the present work), resulting in the set \(\{I_1,I_2,...,I_N\}\). These images are captured from \textit{N} points of view, whose poses are known and stored \(\vec{P_i}=(x_i,y_i,\theta_i),i=1,...,N\). Additionally the room where the picture has been captured is known too, so a set of labels is available: \(\vec{R}_i=(r_i),i=1,...,N\). Each image will be embedded into a single descriptor during the localization, using the proposed architecture, yielding \(\{\vec{f}_1,\vec{f}_2,...,\vec{f}_N\}\). The trajectory followed by the robot includes different rooms with different visual information. In this work, these rooms include a corridor, some offices, a library and a bathroom.

\vspace{0.5cm}
Taking these facts into account, the initial map is composed by the set of images, their poses and the room in which the images are captured \(\{(I_1,\vec{P}_1,r_1),(I_2,\vec{P}_2,r_2),...,(I_N,\vec{P}_N,r_N)\}\). Using this information, some Siamese Neural Networks are trained to address localization.

\subsection{Room Discrimination}

In this subsection an initial task related to localization is evaluated to study whether a Siamese Neural Network is able to distinguish between images captured from the same or from different rooms. For this purpose, the model will be trained and tested with pairs of random images captured from the same and/or different room. 

\subsection{Global Localization}

In this study we consider that a map of the environment is available, as described before. The absolute localization problem is solved by comparing the test image directly with all the images in the map. This comparison is performed using the descriptors \(\vec{f}_i\) associated to each image in the map. The pose of the robot is found as the most similar descriptor contained in that map. The problem is approached with pure visual information and assuming that no information about the previous pose of the robot is available.

% the localization phase can be addressed using different strategies: First, we asses a room retrieval strategy, that tries to identify the room in which the test image was captured

% On the one hand, the hierarchical localization is performed following a room retrieval strategy that tries to identify first the room in which the test image is captured. Next, the most similar pose among the images in that room is retrieved. On the other hand, the absolute localization problem consists in estimating the pose of the robot comparing the test image directly with all the images in the model. The problems are approached with pure visual information and assuming that no information about the previous pose of the robot is available. 

% To solve the localization problem, visual information is described using global appearance approaches. The absolute localization is performed as a pairwise comparison of the visual descriptors. The test image descriptor is compared with the descriptors of the images stored in the model and the most similar pair is selected to estimate the pose of the robot. 

% Following the idea of the hierarchical localization, the task is performed in two steps. Firstly, the test image is compared with a representative image of each room and the most similar representative is selected as the room in which the robot is. After that, in a second step, the most similar image of the model is retrieved using the pairwise comparison, considering only with the images that belong to the previously retrieved room. 

\section{Architecture and Training of the Deep Learning tools}
\label{sec:Architecture}
\begin{figure}[]
\vspace{-0.2cm}
\centering
\includegraphics[width=\linewidth]{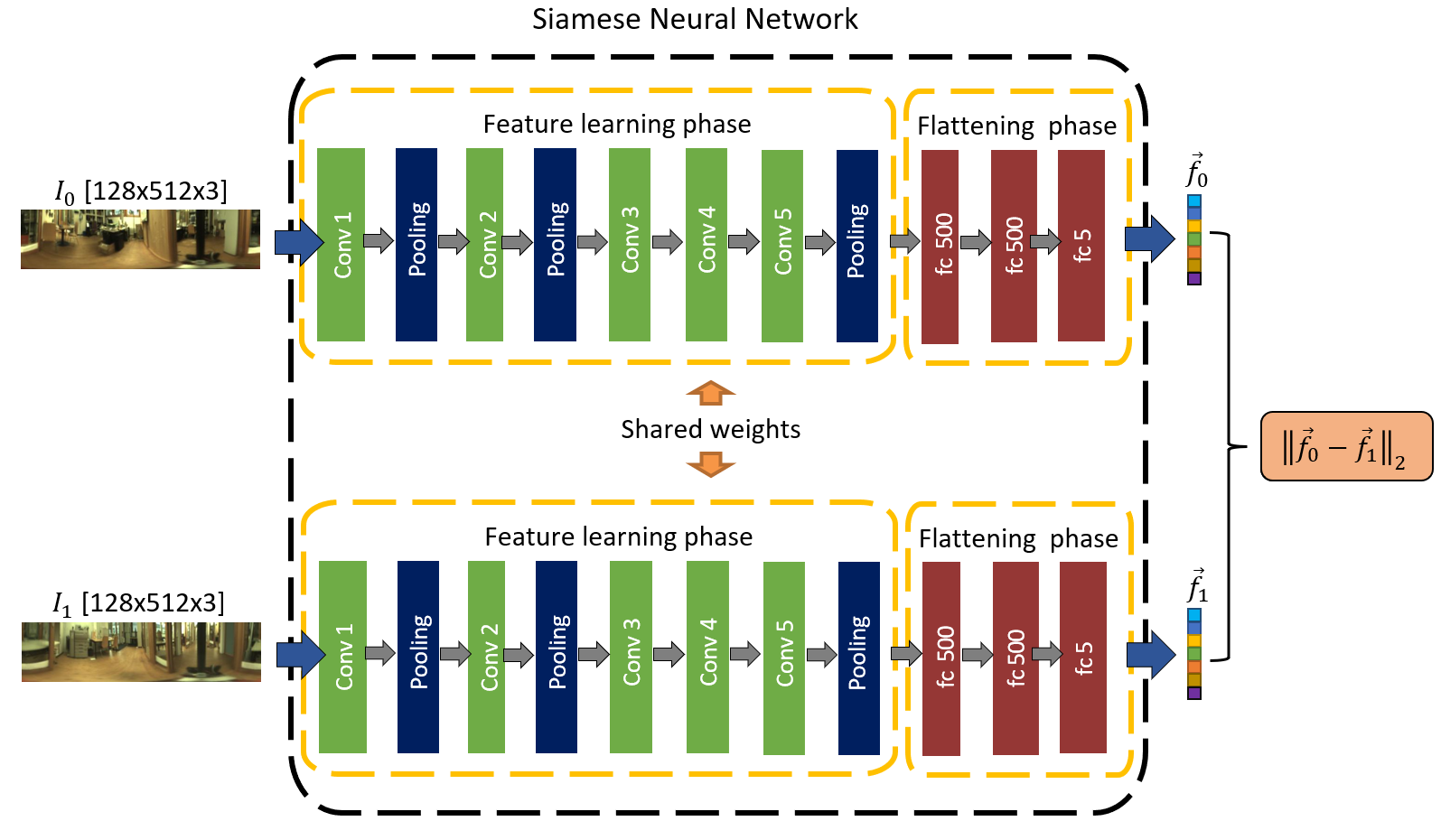}

%   {\epsfig{file = Figures/SiameseArquitecture_completa.PNG, width = 14cm}}
  \caption{Detailed representation of a Siamese Neural Network with AlexNet in the feature learning phase and the flattening phase composed of three fully connected layers.}
  \label{fig:siamese_completa}
  \vspace{-0.1cm}
\end{figure}

%\noindent This 
The structure of a classical CNN used for classification tasks can be split into two different stages \citep{cebollada2019hierarchical}: the feature learning and the classification stages. Features are extracted using several convolutional layers whereas the classification task can be constructed using fully connected layers and a final Softmax function. In our approach, the classification stage is replaced by a flattening phase. In this sense, the feature learning phase outputs a feature map which is flattened to a vector and dimensionally reduced by fully connected layers. This phase permits generating a single description vector per input image. As a result, the model provides two vectors \(\vec{f}_0\) and \(\vec{f}_1\) (one per input image). These descriptors are compared using the Euclidean distance in the comparison phase \((d(\vec{f}_0,\vec{f}_1)=\|\vec{f}_0-\vec{f}_1\|_2)\). This architecture is shown in Figure \ref{fig:siamese_completa}. Therefore, during training, the weights of the networks are updated in order to obtain the optimal global descriptors. After the comparison, the distance between them and the similarity label  \((1:dissimilar, 0:similar)\) are used as data for the loss function. In our case the loss function used is the  Constrastive Loss Function.

% Taking into account the previous idea and the aim of the work in which the siamese network share weights and learns altogether, the feature learning stage consists of two different networks that share their weights and compute the descriptors. Then the descriptors are compared and evaluated in the comparison stage (Figure \ref{fig:siamese_completa}).

\unskip

\begin{equation}
L(\vec{f}_0,\vec{f}_1)=\frac{1}{2}(1-y)d(\vec{f}_0,\vec{f}_1)^2+ y\frac{1}{2}max(\alpha-d(\vec{f}_0,\vec{f}_1),0)^2
\label{eq:ContrastiveLoss}
\end{equation}

Where \(y\) is the similarity label and \(\alpha > 0\) is a margin. The margin defines a radius
around the descriptor so that dissimilar pairs of images contribute to the loss function only if their distance is within this radius \citep{hadsell2006dimensionality}.

\subsection{Parameters and Networks}
\label{subsec:Parameters}
%\noindent This 
 
In this manuscript we compare different networks in the feature learning stage. As inputs to the flattening layers we consider the representation computed in the last convolutional layer of Alexnet \citep{krizhevsky2012imagenet}, DenseNet \citep{he2016deep}, VGG11, VGG13, VGG16 and VGG19 \citep{simonyan2014VVG}. In Table \ref{tab:Layers} the feature extraction layers of those CNNs are presented. Additionally two simple networks created with three conv2d layers are also evaluated (Table \ref{tab:SimpleCNNS}). The ReLU activation layers are not shown for brevity, but they have been used after each conv2d layer. The different feature learning structures are evaluated in the Section \ref{sec:Experiments}.

\begin{table}[!htb]
\caption{Configuration of the Feature Extraction Neural Networks. The ReLU activation layers have been omitted for brevity.}\label{tab:Layers}
\centering
\begin{tabular}{|cccccc|}
\hline
\multicolumn{1}{|c|}{AlexNet}                                                                      & \multicolumn{1}{c|}{DenseNet}       & \multicolumn{1}{c|}{VGG11}                                                            & \multicolumn{1}{c|}{VGG13}                                                           & \multicolumn{1}{c|}{VGG16}                                                                         & VGG19                                                                                       \\ \hline
\multicolumn{6}{|c|}{input (128 x 512 RGB image)}                                                                                                                                                                                                                                                                                                                                                                                                                                                                        \\ \hline
\multicolumn{1}{|c|}{conv2d-64}                                                                    & \multicolumn{1}{c|}{conv2d-112}     & \multicolumn{1}{c|}{conv2d-64}                                                        & \multicolumn{1}{c|}{\begin{tabular}[c]{@{}c@{}}conv2d-64\\ conv2d-64\end{tabular}}   & \multicolumn{1}{c|}{\begin{tabular}[c]{@{}c@{}}conv2d-64\\ conv2d-64\end{tabular}}                 & \begin{tabular}[c]{@{}c@{}}conv2d-64\\ conv2d-64\end{tabular}                               \\ \hline
\multicolumn{1}{|c|}{maxpool}                                                                      & \multicolumn{1}{c|}{maxpool}        & \multicolumn{4}{c|}{maxpool}                                                                                                                                                                                                                                                                                                                                                    \\ \hline
\multicolumn{1}{|c|}{conv2d-192}                                                                   & \multicolumn{1}{c|}{conv2d-56 x 6}  & \multicolumn{1}{c|}{conv2d-128}                                                       & \multicolumn{1}{c|}{\begin{tabular}[c]{@{}c@{}}conv2d-128\\ conv2d-128\end{tabular}} & \multicolumn{1}{c|}{\begin{tabular}[c]{@{}c@{}}conv2d-128\\ conv2d-128\end{tabular}}               & \begin{tabular}[c]{@{}c@{}}conv2d-128\\ conv2d-128\end{tabular}                             \\ \hline
\multicolumn{1}{|c|}{maxpool}                                                                      & \multicolumn{1}{c|}{averagepool}    & \multicolumn{4}{c|}{maxpool}                                                                                                                                                                                                                                                                                                                                                    \\ \hline
\multicolumn{1}{|c|}{\begin{tabular}[c]{@{}c@{}}conv2d-384\\ conv2d-256\\ conv2d-256\end{tabular}} & \multicolumn{1}{c|}{conv2d-28 x 12} & \multicolumn{1}{c|}{\begin{tabular}[c]{@{}c@{}}conv2d-256\\  conv2d-256\end{tabular}} & \multicolumn{1}{c|}{\begin{tabular}[c]{@{}c@{}}conv2d-256\\ conv2d-256\end{tabular}} & \multicolumn{1}{c|}{\begin{tabular}[c]{@{}c@{}}conv2d-256\\ conv2d-256\\ conv2d-256\end{tabular}}  & \begin{tabular}[c]{@{}c@{}}conv2d-256\\  conv2d-256\\ conv2d-256\\  conv2d-256\end{tabular} \\ \hline
\multicolumn{1}{|c|}{\multirow{5}{*}{maxpool}}                                                     & \multicolumn{1}{c|}{averagepool}    & \multicolumn{4}{c|}{maxpool}                                                                                                                                                                                                                                                                                                                                                    \\ \cline{2-6} 
\multicolumn{1}{|c|}{}                                                                             & \multicolumn{1}{c|}{conv2d-14 x 24} & \multicolumn{1}{c|}{\begin{tabular}[c]{@{}c@{}}conv2d-512\\ conv2d-512\end{tabular}}  & \multicolumn{1}{c|}{\begin{tabular}[c]{@{}c@{}}conv2d-512\\ conv2d-512\end{tabular}} & \multicolumn{1}{c|}{\begin{tabular}[c]{@{}c@{}}conv2d-512\\ conv2d-512\\ conv2d-512\end{tabular}}  & \begin{tabular}[c]{@{}c@{}}conv2d-512\\ conv2d-512\\ conv2d-512\\  conv2d-512\end{tabular}  \\ \cline{2-6} 
\multicolumn{1}{|c|}{}                                                                             & \multicolumn{1}{c|}{averagepool}    & \multicolumn{4}{c|}{maxpool}                                                                                                                                                                                                                                                                                                                                                    \\ \cline{2-6} 
\multicolumn{1}{|c|}{}                                                                             & \multicolumn{1}{c|}{conv2d-7 x 16}  & \multicolumn{1}{c|}{\begin{tabular}[c]{@{}c@{}}conv2d-512\\ conv2d-512\end{tabular}}  & \multicolumn{1}{c|}{\begin{tabular}[c]{@{}c@{}}conv2d-512\\ conv2d-512\end{tabular}} & \multicolumn{1}{c|}{\begin{tabular}[c]{@{}c@{}}conv2d-512\\ conv2d-512\\  conv2d-512\end{tabular}} & \begin{tabular}[c]{@{}c@{}}conv2d-512\\ conv2d-512\\ conv2d-512\\ conv2d-512\end{tabular}   \\ \cline{2-6} 
\multicolumn{1}{|c|}{}                                                                             & \multicolumn{1}{c|}{averagepool}    & \multicolumn{4}{c|}{maxpool}                                                                                                                                                                                                                                                                                                                                                    \\ \hline
\multicolumn{6}{|c|}{\textcolor{blue}{fc-500} }                                                                                                                                                                                                                                                                                                                                                                                                                                                                                              \\ \hline
\multicolumn{6}{|c|}{\textcolor{blue}{fc-500}     }                                                                                                                                                                                                                                                                                                                                                                                                                                                                                          \\ \hline
\multicolumn{6}{|c|}{\textcolor{blue}{fc-5}    }                                                                                                                                                                                                                                                                                                                                                                                                                                                                                             \\ \hline
\end{tabular}
\footnotetext{* Blue color layers correspond to the flattening layers.}
\footnotetext{**VGG networks have their Batch Normalize (bn) version where after each conv2d layer a BatchNorm2d layer normalizes the results.}
\end{table}

\begin{table}[!htb]
\caption{Simple Convolutional Neural Networks without pretraining.}\label{tab:SimpleCNNS}
\centering
\begin{tabular}{|cc|}
\hline
\multicolumn{1}{|c|}{Simple 1}                                                                & Simple 2                                                                 \\ \hline
\multicolumn{2}{|c|}{input (128 x 512 RGB image)}                                                                                                                        \\ \hline
\multicolumn{1}{|c|}{\begin{tabular}[c]{@{}c@{}}conv2d-3\\ conv2d-8\\ conv2d-16\end{tabular}} & \begin{tabular}[c]{@{}c@{}}conv2d-3\\ conv2d-16\\ conv2d-32\end{tabular} \\ \hline
\multicolumn{2}{|c|}{maxpool}                                                                                                                                            \\ \hline
\multicolumn{2}{|c|}{\textcolor{blue}{fc-500}   }                                                                                                                                          \\ \hline
\multicolumn{2}{|c|}{\textcolor{blue}{fc-500}  }                                                                                                                                             \\ \hline
\multicolumn{2}{|c|}{\textcolor{blue}{fc-5}  }                                                                                                                                               \\ \hline
\end{tabular}
\footnotetext{* Blue color layers correspond to the flattening layers.}
\end{table}

We have also tested different structures in the flattening phase. As a global baseline three fully connected layers are used, but different versions are considered, with different number of neurons. The different layers used during the evaluation are presented in Table \ref{tab:Layers_Classif}.

\begin{table}[!htb]
\centering
\caption{ Configuration of the flattening phase in our approach.}\label{tab:Layers_Classif}
\begin{tabular}{|c|c|c|}
\hline
\textbf{version 1} & \textbf{version 2} & \textbf{version 3} \\ \hline
fc-500             & fc - 500           & fc - 1000          \\ 
fc - 500           & fc - 100           & fc - 1000          \\ 
fc - 5             & fc - 10            & fc - 10            \\ \hline
\end{tabular}
\end{table}

Other parameters are also tested during the training phase with the aim of obtaining the best Siamese Neural Network for our application. The hyperparameters considered during the evaluation are the following: the batch size (number of samples processed  before the model is updated), the epochs (number of complete passes through the training dataset) and the percentage of images (percentage of training pairs of images from the same or different rooms, so that the network can learn adequately similarities and dissimilarities between rooms). In the experiments, the learning rate is kept constant at 0.001 (rate of change of the model in response to the estimated error) and the momentum is 0.9 (contribution of the parameter update step of the previous iteration upon the current iteration).

\subsection{Data Augmentation}
\label{subsec:DA}

Additionally, a data augmentation technique is proposed as a method to improve the performance of the network. It increases the number of images in the training dataset. Having a larger number of training images reduces the overfitting of the model and boosts its robustness against real operation conditions.  \cite{cabrera2021robust} and \cite{sakkos2019illumination} demonstrated the use of data augmentation in CNNs to improve their effectiveness under changing lighting conditions.

\vspace{0.5cm}

Our proposed data augmentation is focused mainly on such lighting conditions and concentrates on editing local regions by simulating lights, reflections and shadow effects caused by light sources from different angles. Moreover global illumination changes are also taken into account. Other effects not related with the illumination but that can appear when images are captured in real operating conditions are also used.

\vspace{0.5cm}

 \begin{description}
    \item[\textbf{Local effects:}] Light sources that fall on a specific area or the surface of an object are reproduced. We call this local illumination changes since only a small patch of the image is being affected. The shape of different light sources can vary meaningfully.
   
    \vspace{0.5cm}
 
    Circular shapes from light bulbs or square and trapezoid shapes from reflections or windows are common. We edit the intensity of different regions following these shapes to simulate the light source; the pixel intensity is increased to reproduce more bright or it is decreased to simulate a shadow effect. In order to replicate a realistic fading effect, the intensity of brightening/darkening is gradually decreased from the center to the edge as an attenuation of the light. 
    
    \vspace{0.5cm}
    The size of the shapes and the position is selected randomly to simulate the effect in different ways and so does the maximum value to consider different intensities. In our experiments these figures are built with sizes between 15 and 40 pixels, different intensities are applied and the patch is degraded from intensity values +/- 160 or 100 to 5. The effects and shapes are shown in the Figure \ref{fig:Local_efects}.
    
%      \begin{figure}[]
% \centering{

% \subfigure{
% \includegraphics[width=4.8cm]{bombilla_luz.jpg}}\quad
% \subfigure{\includegraphics[width=4.8cm]{bombilla_osc.jpg}}\quad
% \subfigure{\includegraphics[width=4.8cm]{cuadrado.jpg}}\quad
% \subfigure{\includegraphics[width=4.8cm]{trapecio.jpg}}\quad
%  %\hfill
% % \subfigure[Darkness light bulb effect.]{\epsfig{file= Figures/bombilla_osc.jpg,width = 5.8cm}\label{Saa_2}}\quad
% % \subfigure[Brightened square effect.]{\epsfig{file= Figures/cuadrado.jpg,width = 5.8cm}}\quad
% % \subfigure[Brightened trapezoid effect.]{\epsfig{file= Figures/trapecio.jpg,width = 5.8cm}}\quad
% %  %\hfill
%     \caption{Individual local effects for data augmentation based on illumination.}  
%       \label{fig:Local_efects}
% }
% \end{figure}

      \begin{figure}[H]
\centering{

\subfigure[Brightened light bulb effect.]{\epsfig{file= 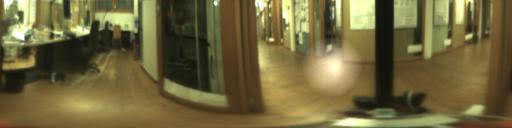,width = 5.5cm}}\quad
 %\hfill
\subfigure[Darkness light bulb effect.]{\epsfig{file= 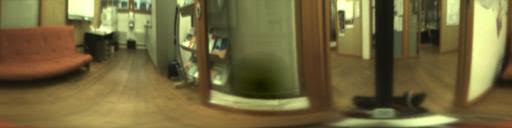,width = 5.5cm}\label{Saa_2}}\quad
\subfigure[Brightened square effect.]{\epsfig{file= 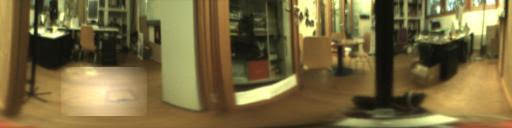,width = 5.5cm}}\quad
\subfigure[Brightened trapezoid effect.]{\epsfig{file= 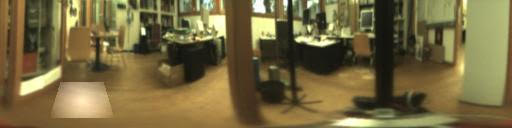,width = 5.5cm}}\quad
 %\hfill
    \caption{Individual local effects for data augmentation based on illumination.}  
      \label{fig:Local_efects}
}
\end{figure}

    \item[\textbf{Global illumination:}] Global illumination variations can occur in some cases. To model such illumination changes, we need to alter pixels across the whole image, rather than in a small region. A constant value \(c\) is added to all the pixels to model a global brightness effect on the image or it is subtracted to simulate a global darkness. The value of \(c\) varies from 35 to 75 in this work. Figures \ref{subf:light} and \ref{subf:dark} show the effect.
    \vspace{0.5cm}
    \item[\textbf{Sharpness/Blurring:}] Finding sharper borders among diverse objects will contribute to provide a better separation among them and between foreground and background. In contrast, blurring effects are caused by low illumination and movements of the camera, which are common in mobile robotics. Both effects are incorporated in the data augmentation. They can be observed in Figures \ref{subf:Sharpness} and \ref{subf:Blurring}. Both can be achieved by a convolution operation using the following masks.
    \begin{table}[!htb]
    \centering
\begin{tabular}{cccc}
 \textbf{Sharpness effect} & & & \textbf{Blurring effect} \\
 \(m_{sh}=\begin{bmatrix}{0}&{-1}&{0}\\{-1}&{5}&{-1}\\{0}&{-1}&{0}\end{bmatrix}\)& & &\(m_{bl}=\frac{1}{25}\begin{bmatrix}{1}&{1}&{1}&{1}&{1}\\{1}&{1}&{1}&{1}&{1}\\{1}&{1}&{1}&{1}&{1}\\{1}&{1}&{1}&{1}&{1}\\{1}&{1}&{1}&{1}&{1}\end{bmatrix}\)
\end{tabular}
\end{table}

    \item[\textbf{Contrast variation:}] The contrast of the image plays an important role in highlighting different objects in the scene. Low contrast images usually look softer and with less shadows and reflections. The effect is proposed for this data augmentation to improve the robustness of the framework. The contrast is modified following the next equation: 
    \[I_s=64+c*(I-64)\]
    
    where \(I_s\) is the resulting image,  \(I\) the original image and  \(c\) is the contrast factor. For  \(c>1\) the contrast increases and \(c<1\) decreases the contrast. 
    Additionally, an equalization of the image is also added to the data augmentation set. It evenly distributes the histogram values, which permits obtaining a new contrast augmentation effect. Figure \ref{subf:Contrast} shows this effect.
    
    \vspace{0.5cm}
    \item[\textbf{Saturation changes:}] The colour saturation of the image deals with the intensity of the colour. The less saturation, the less colourful the image is, even it can resemble a grey-scale image if the saturation is very low. In contrast, more vivid colours are obtained when the colour saturation is high. It can simulate situations when illumination changes significantly. 
    The colour saturation can be edited by converting the RGB image to HSV, after that, it is possible to directly change the saturation channel by multiplying it by a constant factor $c$. If the saturation  attribute is multiplied by \(c>1\) the colours become more saturated and by \(c<1\) the colour saturation decreases. The effect can be seen in Figure \ref{subf:Saturation}.
    
    \vspace{0.5cm}
    \item[\textbf{Rotation:}] The original image covers 360 degrees around the robot. For that reason the image can be rotated without losing any piece of information. This effect will simulate the situation in which the robot is in the same position but the orientation is different. Moreover, having a training dataset containing this type of effect is expected to provide the Neural Network with rotation invariance. Figure \ref{subf:Rotation} shows a rotation effect of 115 degrees. Random rotations between 10 and 350 degrees are applied to the training images.
    
        \begin{figure}[H]
\centering{
\subfigure[Original image.]{

\epsfig{file= 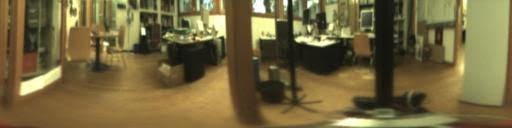,width = 5.5cm}}\quad %\hfill
\subfigure[Global light illumination.]{\epsfig{file= 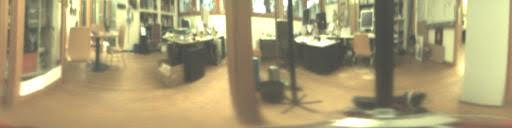,width = 5.5cm}\label{subf:light}}\quad
\subfigure[Global dark illumination.]{\epsfig{file= 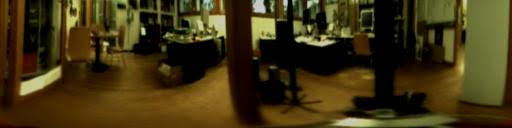,width = 5.5cm}\label{subf:dark}}\quad
\subfigure[Sharpness effect.]{\epsfig{file= 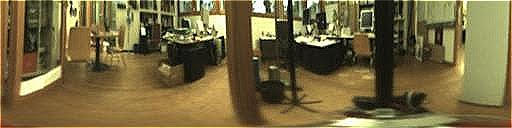,width = 5.5cm}\label{subf:Sharpness}}\quad
\subfigure[Blurring effect.]{\epsfig{file= 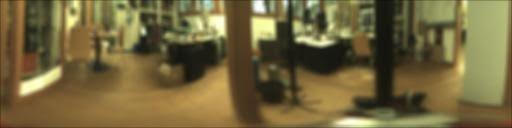,width = 5.5cm}\label{subf:Blurring}}\quad
 %\hfill
\subfigure[Contrast variation effect.]{\epsfig{file= 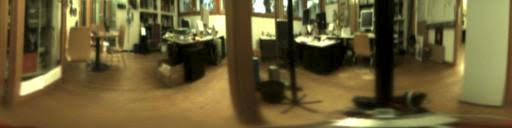,width = 5.5cm}\label{subf:Contrast}}\quad
\subfigure[Saturation changes effect.]{\epsfig{file= 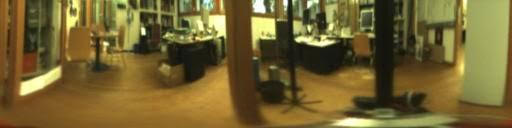,width = 5.5cm}\label{subf:Saturation}}\quad
\subfigure[Rotation effect.]{\epsfig{file= 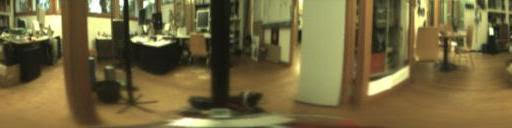,width = 5.5cm}\label{subf:Rotation}}\quad
 %\hfill
    \caption{Global effects for data augmentation. }  
      \label{fig:Global_efects}
}
\end{figure}

    \item[\textbf{Combined changes:}] Additionally some effects are combined to obtain a larger data augmentation, but not all the effects are combined together. Global illumination and a single local effect are combined in all the possible variations, e.g. global darkness is combined with a brightening circle shape effect, global brightness is combined with a brightening trapezoidal effect, etc. Additionally, the local effects are also combined. The circle shape effect is combined with the square effect, the trapezoidal effect or another circle shape effect, the combinations can be brightened+brightened, brightened+darkness and darkness+darkness; the circle shape effect is also combined with other two circle shape effects, obtaining an image with three light bulb effects. Finally, the rotation effect is individually combined with all the effects and the combinations described above.
    
\end{description}

\subsection{Training and testing the Siamese Neural Network}
\label{sec:Solving}

% Training a network model from scratch is tough, requires experience with neural network architectures, a significant amount of data and a huge computational time. Due to these reasons, a CNN is adapted to each siamese branch in order to obtain a Siamese Neural Network which is retrained to solve the new task by a transfer learning technique.
As presented in Section \ref{subsec:Parameters}, different CNNs architectures can be used as the base of Siamese Neural Networks. Initially, we start from pretrained networks with known weights and biases. Then, we retrain the network to fit it to our application. This transfer learning technique is well-known and has previously been used in mobile robotics \citep{cabrera2021robust,cebollada2020deep}.

\vspace{0.5cm}

The following Subsection \ref{subsec:RoomRetrieval} will address an initial task which consists in training and evaluating the capability of a Siamese Neural Network to identify whether two images were captured from the same or different rooms. Finally, in Subsection \ref{subsec:AbsoluteLoc} we will detail the characteristics of the training and test to address the absolute localization problem with siamese architectures. Emphasis will be placed on the labelling required to perform the desired task.

\subsubsection{Room Discrimination}
%\noindent This 
\label{subsec:RoomRetrieval}
The main goal of this task is to evaluate whether a Siamese Neural Network is capable of determining if two images belong to the same or different room. It is an important capability to perform localization tasks.  

% It is an important problem, since the first step of hierarchical localization consists in comparing the test image with a representative of each room. In this task, we consider any image of the model as representative of the room. 

\vspace{0.5cm}
The \textit{training phase} is performed by feeding the network with pairs of images. These pairs are labelled with 0 if they have been captured from the same room and 1 if not. The ratio same/different room pairs is varied in the \textit{training phase} to study its influence. 

% During this phase 8486 images captured in cloudy, sunny and night conditions are used. 

 \begin{figure}[H]
 \vspace{-0.2cm}
  \centering
  \includegraphics[width=\linewidth]{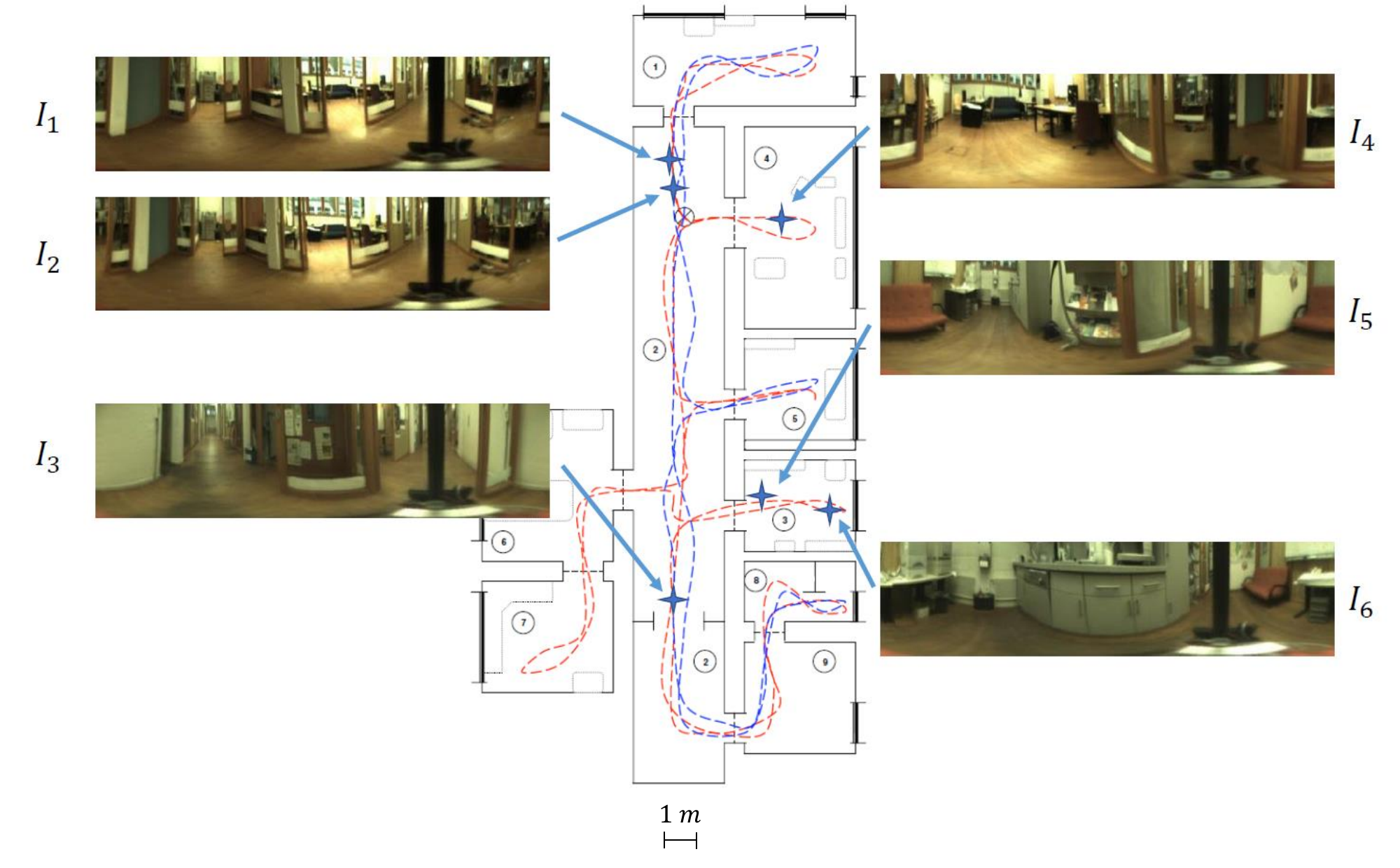}
%   {\epsfig{file = Figures/AbsoluteLoc.pdf, width = 14cm}}
  \caption{ Example of different trajectories of the robot.} 
  \label{fig:AbsoluteLoc}
  \vspace{-0.1cm}
\end{figure}
\unskip

% \begin{table}[]
% \centering
% \caption{Example pairs and its label value for the room retrieval task.}\label{tab:Label_room}
% \begin{tabular}{|c|c|}
% \hline
% \textbf{Pair} & \textbf{Label Value} \\ \hline
%       \(I_1-I_2\)                &            0           \\
%       \(I_1-I_3\)     &            0           \\
%       \(I_1-I_4\)     &            1           \\
%       \(I_1-I_5\)     &            1           \\
%       \(I_4-I_5\)     &            1           \\
%       \(I_5-I_6\)     &            0           \\ \hline
% \end{tabular}
% \end{table}

\vspace{0.5cm}
During the \textit{test phase}, pairs of images are fed into the network. At the output, the network labels them with a number between 0 and 1; if the result is under 0.5 we interpret that the images have been captured from the same room. On the contrary, the images belong to different rooms. The images used to test the network are different from the training images, they are captured in the same building but in different times, in a variety of lighting conditions. Also the trajectory followed by the robot to capture the test images is similar to the one used to capture the training images, but the images are captured from different robot poses (Figure \ref{fig:AbsoluteLoc}).

\subsubsection{Global localization}
%\noindent This 
\label{subsec:AbsoluteLoc}

The global localization problem considers the estimation of the robot pose within the whole floor of the building. For this purpose, a Siamese Neural Network is trained. The \textit{training} is carried out with image pairs labelled with the following equation:
 \begin{equation}
 Label(I_i, I_j)= \left\{ \begin{array}{lcc}
             \frac{\|\vec{p}_i-\vec{p}_j\|_2}{K_b}  & \mathrm{if} ~I_{i}~ \mathrm{and} ~I_{j}~ \mathrm{belong~ to ~the~ same ~room} \\
             1 &  \mathrm{otherwise} 
             
             \end{array}
   \right.
  \label{eqn:labelling_loc_global}
 \end{equation}

% If \(I_{i}\) and \(I_{j}\) belong to the same room:
% \[ Label(I_i, I_j) = \frac{\|\vec{p}_i-\vec{p}_j\|_2}{K_b}  \]

% If \(I_{j}\) and \(I_{k}\) belong to the different rooms:
% \[ Label(I_{j}, I_{k}) = 1  \] 
% Where \( I_{j}\) and \(I_{k}\) are two images poses captured in the same room. 
 \noindent where $I_i$ and $I_j$ are two images and \( \vec{p}_i\) and \(\vec{p}_j\) are their corresponding positions (coordinates of the capture points). This constitutes a normalized Euclidean distance between the capture points. $K_b$ corresponds to the maximum distance between two images in the building. Table \ref{tab:Label_loc} shows different examples according to Figure \ref{fig:AbsoluteLoc}.

 \begin{table}[!htb]
\centering
\caption{Example pairs and its label value for the absolute localization task. The labels of the images are shown in Figure \ref{fig:AbsoluteLoc}.}\label{tab:Label_loc}
\begin{tabular}{|c|c|c|}
\hline
\textbf{Pair} & \textbf{Euclidean distance (m)} & \textbf{Label Value}\\ \hline
       \(I_1-I_2\)     &    0.33    &        \(\frac{0.33}{18.99}=0.017\) \\
       \(I_1-I_3\)     &    12.82   &        \(\frac{12.82}{18.99}=0.675\) \\
       \(I_1-I_4\)     &    -       &            1           \\
       \(I_1-I_5\)     &    -       &            1           \\
       \(I_4-I_5\)     &    -       &            1           \\
       \(I_5-I_6\)     &    2.48    &        \(\frac{2.48}{18.99}=0.131\)  \\ \hline
\end{tabular}
\begin{description}
\item Where: \\18.99 m is the maximum distance between two images in the target environment
\end{description}
\end{table}

Once the network has been trained, the test is performed by using the map which is composed by the set of image descriptors and their positions \(\{(\vec{f}_1,\vec{p}_1),(\vec{f}_2,\vec{p}_2),...,(\vec{f}_N,\vec{p}_N)\}\). Each descriptor has been calculated by the trained Siamese Neural Network. The absolute localization is performed as a pairwise comparison between image descriptors. Given a test image $I_t$, the Siamese Neural Network outputs its corresponding descriptor $\vec{f}_t$. Finally, the position of the robot is estimated by selecting the pose associated to the descriptor in the map that minimizes the distance $\|\vec{f}_t-\vec{f}_i\|_2$, with $i = 1,\ldots, N$.

%  After the network has been trained the test is performed. The robot takes a new image at instant \textit{t} from an unknown pose. The new image \(f_t\) is compared to all the images in the base model, each comparison calculates an associated similarity value between images. Once the new image is compared to the images in the base trajectory the comparison with the lowest value is selected as the more similar image and the pose of the robot is estimated. \textcolor{red}{Explicar mejor. A partir de la imagen de test se genera un descriptor el cual se compara con los previamente calculados dEL MAPA mediante un calculo de la distancia euclidea entre pares de descriptor. FUSIONAR CON LA SECCION 6.3 }
 
%  Metric information is used in the training phase and as a ground truth, but not used in the \textit{test phase}, where only visual information is used. 
 
%  The \textit{training phase} has been performed with three different perspectives. Firstly, following the room retrieval idea, the training has been prepared using images from the three illumination variations (cloudy, night and sunny) so the same 8486 images used for the room retrieval problem are used. Secondly, it is also studied the case in which only images from the cloudy environment are taken, for improving this circumstance the 2778 images are used with a data augmentation process explained in the subsection \ref{subsec:DA}. On the other hand, the \textit{test phase} is performed with images captured in three illumination conditions.  \textcolor{red}{quitar}

\section{Experiments}
\label{sec:Experiments}
The set of experiments is designed to test the performance of the Siamese Neural Network as global descriptor generator to tackle the room discrimination and global localization task as explained in subsections \ref{subsec:RoomRetrieval} and \ref{subsec:AbsoluteLoc}. Furthermore, the training and test datasets used in this research are presented in the following subsection.

\subsection{Training and test datasets}\label{sec:dataset}

The images used in the experiments are obtained from an indoor dataset \citep{pronobis2009COLD}. This database was captured by an omnidirectional vision sensor mounted on a mobile robot which followed different trajectories that visited 9 different rooms. A variety of lighting conditions was considered to capture the sets of images.

\vspace{0.5cm}

Table \ref{tab:tabla_datos} shows the number of images per room for each of the datasets used in this research. Two training sets are considered: training set 1 consists of 8486 images captured under cloudy, sunny and night illumination conditions (\textit{COLD-Freiburg Part A Path 2 Cloudy 3, Freiburg Part A Path 2 Night 1, Freiburg Part A Path 2 Sunny 3}). Training set 2 has been obtained by applying a data augmentation to the cloudy sequence of training set 1, thus generating 977856 images. With respect to the test sets, four different sets are considered: test set 1 consists of 2595 images under cloudy lighting condition (\textit{COLD-Freiburg Part A Path 2 Cloudy 2}), test set 2 contains images captured under night lighting condition and consists of 2707 images (\textit{COLD-Freiburg Part A Path 2 Night 2}), test set 3 consists of 2114 images under sunny lighting condition (\textit{COLD-Freiburg Part A Path 2 Sunny 2}) and test set 4 is composed of all the images in the previous test sets. It should be noted that the images in the test sets are different, in all cases, from the images that constitute the training sets. Finally, the visual map has been obtained after sampling the path under the cloudy lighting condition of the training set 1, obtaining a total of 556 images.

\vspace{0.5cm}

In this way, the training sets will be used to carry out the training of the Siamese Neural Networks, and the test sets will evaluate the performance of the networks under the three lighting conditions. The visual model is the map available for the robot to carry out the localization, so it will be used in the testing phase of the global localization.

\begin{table}[!htb]
\caption{Summary of the training and test datasets. This table shows the number of images per room and the total of images of each dataset.}
\label{tab:tabla_datos}
\begin{tabular}{|c|c|c|c|c|c|c|c|}
\hline
\textbf{Room}   & \textbf{\begin{tabular}[c]{@{}c@{}}Training\\ dataset 1\end{tabular}} & \textbf{\begin{tabular}[c]{@{}c@{}}Training\\ dataset 2\end{tabular}} & \textbf{\begin{tabular}[c]{@{}c@{}}Test\\ dataset 1\end{tabular}} & \textbf{\begin{tabular}[c]{@{}c@{}}Test\\ dataset 2\end{tabular}} & \textbf{\begin{tabular}[c]{@{}c@{}}Test\\ dataset 3\end{tabular}} & \textbf{\begin{tabular}[c]{@{}c@{}}Test\\ dataset 4\end{tabular}} & \textbf{\begin{tabular}[c]{@{}c@{}}Visual\\ Map\end{tabular}} \\ \hline
\textbf{1P0-A}  & 518                                                                   & 76,736                                                                & 218                                                               & 168                                                               & 123                                                               & 509                                                               & 44                                                            \\ \hline
\textbf{2P01-A} & 694                                                                   & 82,016                                                                & 233                                                               & 215                                                               & 187                                                               & 635                                                               & 46                                                            \\ \hline
\textbf{2P02-A} & 428                                                                   & 55,616                                                                & 158                                                               & 168                                                               & 109                                                               & 435                                                               & 31                                                            \\ \hline
\textbf{CR-A}   & 3,258                                                                  & 416,416                                                               & 1,183                                                              & 1,114                                                              & 793                                                               & 3,090                                                              & 238                                                           \\ \hline
\textbf{KT-A}   & 674                                                                   & 80,608                                                                & 229                                                               & 270                                                               & 213                                                               & 712                                                               & 46                                                            \\ \hline
\textbf{LO-A}   & 395                                                                   & 46,464                                                                & 132                                                               & 121                                                               & 102                                                               & 355                                                               & 26                                                            \\ \hline
\textbf{PA-A}   & 804                                                                   & 99,968                                                                & 284                                                               & 241                                                               & 191                                                               & 716                                                               & 57                                                            \\ \hline
\textbf{ST-A}   & 495                                                                   & 53,152                                                                & 151                                                               & 198                                                               & 180                                                               & 529                                                               & 30                                                            \\ \hline
\textbf{TL-A}   & 619                                                                   & 66,880                                                                & 190                                                               & 212                                                               & 216                                                               & 618                                                               & 38                                                            \\ \hline
 
\textbf{Total:} & 8,486                                                                 & 977,856                                                               & 2,595                                                             & 2,707                                                             & 2,114                                                             & 7,416                                                              & 556                                                           \\ \hline
\end{tabular}
\end{table}
% \begin{table}[]

% \caption{Summary of the training and test datasets. DEcir que se muestra el numero d eimaggens por habitacion}
% \label{tab:tabla_datos}
% \small
% \begin{tabular}{|l|l|l|l|l|l|l|l|l|l|}
% \hline
% \textbf{Room} &
%   \textbf{1P0-A} &
%   \textbf{2P01-A} &
%   \textbf{2P02-A} &
%   \textbf{CR-A} &
%   \textbf{KT-A} &
%   \textbf{LO-A} &
%   \textbf{PA-A} &
%   \textbf{ST-A} &
%   \textbf{TL-A} \\ \hline
% \textbf{\begin{tabular}[c]{@{}l@{}}Training \\ dataset 1\end{tabular}} & 518  & 694  & 428  & 3258  & 674  & 395  & 804  & 495 &  619  \\
% \hline

% \textbf{\begin{tabular}[c]{@{}l@{}}Training \\ dataset 2\end{tabular}} &
%   76736 &
%   82016 &
%   55616 &
%   416416 &
%   80608 &
%   46464 &
%   99968 &
%   53152 &
%   66880 \\ \hline
% \textbf{\begin{tabular}[c]{@{}l@{}}Test \\ dataset 1\end{tabular}}    & 218 & 233 & 158 & 1183 & 229 & 132 & 284 & 151 & 190 \\ \hline
% \textbf{\begin{tabular}[c]{@{}l@{}}Test \\ dataset 2\end{tabular}}    & 168 & 215 & 168 & 1114 & 270 & 121 & 241 & 198 & 212 \\ \hline
% \textbf{\begin{tabular}[c]{@{}l@{}}Test \\ dataset 3\end{tabular}}    & 123 & 187 & 109 & 793 & 213 & 102 & 191 & 180 & 216 \\ \hline
% \textbf{\begin{tabular}[c]{@{}l@{}}Test \\ dataset 4\end{tabular}}    & 509 & 635 & 435 & 3090 & 712 & 355 & 716 & 529 & 618 \\ \hline
% \textbf{\begin{tabular}[c]{@{}l@{}}Visual \\ map\end{tabular}} & 44  & 46  & 31  & 238  & 46  & 26  & 57  & 30  & 38  \\ \hline
% \end{tabular}

% \end{table}

% These rooms are presented in Table \ref{tab:Rooms}, as well as the maximum distance between images taken at the same room. 

\subsection{Room Discrimination}
In this subsection we assess the ability of the network to predict whether two images are taken from the same room. The effectiveness of the Siamese Neural Network is calculated by comparing pairs of images and checking their label. The results are expressed in percentage of accuracy. Several experiments have been conducted while varying different parameters: the feature extraction architecture, the flattening layers and the percentage of similar/dissimilar images. As common parameters, we train the network using 8486 pairs of images per epoch from the training dataset 1 and we use the Stochastic Gradient Descent (SGD) optimiser, with a learning rate of 0.001 and momentum of 0.9. Moreover, we test the network with 7000 pairs of images extracted from the test dataset 4.

%\begin{enumerate}%[(a)]    %items (quierop ponerlos con letras)
\subsubsection{Influence of the architecture on the feature extraction process}

% \begin{description}
%     \item[Feature Extraction Layers Election:] \hfill \\
    In this subsection we compare different models in the feature extraction stage of a Siamese Neural Network. The different models used can be observed in Table \ref{tab:Layers}. The training has been performed using a batch size of 256 and 5 epochs. During training, the dataloader presents a 50\% of images from the same room and a 50\% of images from the different rooms. During these experiments, the flattening is performed with 3 fully connected layers composed by 500-500-5 neurons in each. 
    
\vspace{0.5cm}
Results are presented in Table \ref{tab:AccuRedes} in terms of global accuracy. Additionally, the test accuracy for the same and different room predictions is also presented. The table shows that the best networks are VGG13 and VGG16. They obtain the best accuracy for predicting pairs of images in the same room (99.44\% and 99.47\% respectively). In addition, VGG13 and VGG16 present the best accuracy predicting if two images are taken from different rooms (79.86\% and 78.91\%). Moreover, the `Simple 1' and `Simple 2' networks obtain considerably good results using only three convolutional layers. Finally, in general terms, it can be observed that all the architectures perform better in predicting whether two images belong to the same room. For this reason, we consider below the possibility of varying the percentage of images of each category in the training phase.

%HASTA AQUI

% when the network predicts images that are at the same room, where rarely it fails, than when the net predict pairs of images taken at different stances, when none of the configurations superior to the 80\% of accuracy. For this reason, next we assess the net using different percentage of images, additionally other parameters like the number of epochs and the batch size are also studied. 

    \begin{table}[!htb]
    \centering
\caption{Accuracy using different Feature Extraction Neural Networks.}\label{tab:AccuRedes}
\begin{tabular}{|l|c|c|c|c|}
\hline
\multicolumn{1}{|c|}{\textbf{Network}}&  \multicolumn{1}{|l|}{\textbf{\begin{tabular}[c]{@{}c@{}}Global Test \\ Accuracy\end{tabular}}} & \multicolumn{1}{|l|}{\textbf{\begin{tabular}[c]{@{}c@{}}Same Room \\ Accuracy\end{tabular}}} & \multicolumn{1}{|l|}{\textbf{\begin{tabular}[c]{@{}c@{}}Different Room \\ Accuracy\end{tabular}}} \\ \hline
Simple 1 & \textcolor{red}{84.59\%}  & 98.16\%  & \textbf{\textcolor{red}{71.03\%}} \\
Simple 2 & 86.45\% &  98.87\% & 74.06\% \\
Alexnet & 86.10\%  & 98.78\%  &  \textcolor{red}{73.41\%}\\
Densenet&  86.06\% & 97.61\% & 74.52\%\\
VGG11 & 87.43\% & 99.08\% & 75.78\% \\
VGG11bn & 87.51\% & 97.49\% & 77.53\% \\
VGG13 & \textbf{\textcolor{green}{89.65\%}} & \textcolor{green}{99.44\%}  & \textbf{\textcolor{green}{79.86\%}} \\
VGG13bn & 88.52\% & 98.26\% & 78.77\% \\
VGG16 & \textcolor{green}{89.19\%} & \textbf{\textcolor{green}{99.47\%}} & \textcolor{green}{78.91\%} \\
VGG16bn & \textbf{\textcolor{red}{82.04\%}} & \textbf{\textcolor{red}{92.68\%}} & 73.39\% \\
VGG19 & 89.17\% & 99.30\%  & 79.04\% \\
VGG19bn & 86.58\% & \textcolor{red}{95.52\%}  & 77.64\% \\\hline
\end{tabular}
\end{table}

\subsubsection{Influence of the training parameters}

    % \item[Percentage of images, batch size and number of epoch:]\hfill \\
In the light of the previous results, next, different training parameters are evaluated. As we explain in the previous subsection, the ratio of training pairs of images in each category is expected to have a substantial influence upon the results. In consequence, we propose to change the percentage of pairs of images at the training phase. The percentage of images taken from the same and different rooms varies from 5\% to 40\% and from 95\% to 60\% respectively. For brevity, we only show the results obtained with VGG13, VGG16 and AlexNet networks. The rest of the training parameters is tuned as before, using 256 as batch size and a flattening phase with three fully connected layers composed by 500, 500 and 5 neurons. The results are presented in Tables \ref{tab:AccuEpochVGG13}, \ref{tab:AccuEpochVGG16} and \ref{tab:AccuEpochAlexNet}. They show a correlation between the percentage of images of same/different room and its respective accuracy, i. e., when the percentage of pairs of images in the same room increases, its associated accuracy also does and a similar phenomenon occurs with the different room category. 
    \begin{table}[!htb]
    \centering
\caption{Accuracy of \textbf{VGG13}. The table presents a variation in the total number of images and in the same-different ratios of training images.}\label{tab:AccuEpochVGG13}
\begin{tabular}{|c|c|c|c|c|c|c|c|}
\hline
\textbf{Epoch} & \textbf{\begin{tabular}[c]{@{}c@{}}Percentage of\\ Training Images\\  (same-different)\end{tabular}} & \textbf{\begin{tabular}[c]{@{}c@{}}Number of \\ Training Images\\ (same-different)\end{tabular}} & \textbf{\begin{tabular}[c]{@{}c@{}}Global \\ Accuracy\end{tabular}} & \textbf{\begin{tabular}[c]{@{}c@{}}Same Room \\ Accuracy\end{tabular}} & \textbf{\begin{tabular}[c]{@{}c@{}}Different Room \\ Accuracy\end{tabular}} \\ \hline
7& 5\%-95\% & 3,046-57,882 & \textcolor{red}{89.88\%} & \textcolor{red}{92.03\%} & 87.73\% \\ \hline
9& 5\%-95\% & 3,917-74,419 & 91.89\% & 92.51\% & 91.27\% \\ \hline
11& 5\%-95\% & 4,787-90,957 & 92.20\% & 92,71\% & \textcolor{green}{91,70\%} \\ \hline
7& 10\%-90\% & 6,093-54,835 & 92.72\% & 98.13\% & 87.30\% \\ \hline
9& 10\%-90\% & 7,834-70,502 & 94.76\%& 98.69\% & 90.82\%\\ \hline
11& 10\%-90\% & 9,574-86,170 & \textcolor{green}{95.08\%} & 98.90\%  & 91.25\%\\ \hline
7& 25\%-75\% & 15,232-45,696 & 93.10\% & 99.09\% & \textcolor{red}{87,12\%}\\ \hline
9& 25\%-75\% & 19,584-58,752 & 93.46\% & 99.06\%& 87.86\%\\ \hline
11& 25\%-75\% & 23,936-71,808 & 93.53\% & \textcolor{green}{99.21\%} & 87.85\%\\ \hline
\end{tabular}
\end{table}

    \begin{table}[!htb]
    \centering
\caption{Accuracy of \textbf{VGG16}. The table presents a variation in the total number of images and in the same-different ratios of training images.}\label{tab:AccuEpochVGG16}
\begin{tabular}{|c|c|c|c|c|c|c|c|}
\hline
\textbf{Epoch} & \textbf{\begin{tabular}[c]{@{}c@{}}Percentage of\\ Training Images\\  (same-different)\end{tabular}} & \textbf{\begin{tabular}[c]{@{}c@{}}Number of \\ Training Images\\ (same-different)\end{tabular}} & \textbf{\begin{tabular}[c]{@{}c@{}}Global \\ Accuracy\end{tabular}} & \textbf{\begin{tabular}[c]{@{}c@{}}Same Room \\ Accuracy\end{tabular}} & \textbf{\begin{tabular}[c]{@{}c@{}}Different Room \\ Accuracy\end{tabular}} \\ \hline
7& 5\%-95\% & 3,046-57,882 & 94.35\%& \textcolor{red}{96.47\%}& 92.23\% \\ \hline
9& 5\%-95\% & 3,917-74,419 & \textcolor{green}{94.94\%}& 96.48\%& \textcolor{green}{93.39\%}\\ \hline
11& 5\%-95\% & 4,787-90,957 & 94.24\%& 97.77\%& 90.72\%\\ \hline
7& 10\%-90\% & 6,093-54,835 & 93.04\%& 97.16\%& 88.92\%\\ \hline
9& 10\%-90\% & 7,834-70,502 & 94.26\%& 97.18\%&91.35\%\\ \hline
11& 10\%-90\% & 9,574-86,170 & 93.59\%& 97.96\%& 89.22\%\\ \hline
7& 25\%-75\% & 15,232-45,696 & 92.46\%& 99.21\%& 85.71\% \\ \hline
9& 25\%-75\% & 19,584-58,752 & 92.28\%& 99.30\%&85.25\%\\ \hline
11& 25\%-75\% & 23,936-71,808 & \textcolor{red}{91.78\%}& 98.81\%& \textcolor{red}{84.74\%}\\ \hline
7& 40\%-60\% & 24,371-36,557 & 92.95\%& 99.38\%& 86.52\%\\ \hline
9& 40\%-60\% & 31,334-47,002 & 92.72\%& 99.48\%& 85.95\%\\ \hline
11& 40\%-60\% & 38,298-57,446 & 93.28\%& \textcolor{green}{99.50\%}& 87.05\%\\ \hline
\end{tabular}
\end{table}

    \begin{table}[!htb]
    \centering
\caption{Accuracy of \textbf{AlexNet}. The table presents a variation in the total number of images and in the same-different ratios of training images.}\label{tab:AccuEpochAlexNet}
\begin{tabular}{|c|c|c|c|c|c|c|c|}
\hline
\textbf{Epoch} & \textbf{\begin{tabular}[c]{@{}c@{}}Percentage of\\ Training Images\\  (same-different)\end{tabular}} & \textbf{\begin{tabular}[c]{@{}c@{}}Number of \\ Training Images\\ (same-different)\end{tabular}} &  \textbf{\begin{tabular}[c]{@{}c@{}}Global \\ Accuracy\end{tabular}} & \textbf{\begin{tabular}[c]{@{}c@{}}Same Room \\ Accuracy\end{tabular}} & \textbf{\begin{tabular}[c]{@{}c@{}}Different Room \\ Accuracy\end{tabular}} \\ \hline
7& 5\%-95\%& 3,046-57,882 & 92.36\%& \textcolor{red}{90.11\%}& \textcolor{green}{94.60\%} \\ \hline
11& 5\%-95\%& 4,787-90,957 & 93.58\%& 94.08\%& 93.07\% \\ \hline
14& 5\%-95\%& 6,093-115,763 & \textcolor{green}{93.68\%}& 94.14\%& 93.22\% \\ \hline
7&  10\%-90\% & 6,093-54,835 & 92.05\%& 94.65\%& 89.44\%\\ \hline
11& 10\%-90\%& 9,574-86,170 & 93.41\%& 96.84\%& 89.98\%\\ \hline
14& 10\%-90\%& 12,186-109,670 & 93.01\%& 97.19\%& 88.82\%\\ \hline
7& 25\%-75\%& 15,232-45,696 & 90.91& 97.54\%& 84.28\%\\ \hline
11& 25\%-75\%& 23,936-71,808 & 91.16\% & 98.92\%& 83.39\%\\ \hline
14& 25\%-75\%& 30,464-91,392 & 90.59\%& 98.28\%& 82.19\%\\ \hline
7& 40\%-60\%& 24,371-36,557 & \textcolor{red}{88.33\%}& 98.80\%& 77.85\%\\ \hline
11& 40\%-60\%& 38,298-57,446 & 88.65\%& 99.07\%& 78.23\%\\ \hline
14& 40\%-60\%& 48,742-73,114 & 88.54\% & \textcolor{green}{99.25\%}& \textcolor{red}{77.82\%}\\ \hline
\end{tabular}
\end{table}

\vspace{0.5cm}

Until this moment, all the experiments have been performed using 256 as batch size, but other values have been tested in order to check the best configuration. Tables \ref{tab:AccuBSVGG16} and \ref{tab:AccuBSAlexNet} show the accuracy using different batch sizes. They show that the global accuracy increases when the batch size is lower.

    \begin{table}[!htb]
    \centering
\caption{Accuracy using \textbf{VGG16} and different batch sizes.}\label{tab:AccuBSVGG16}
\begin{tabular}{|c|c|c|c|c|c|c|c|}
\hline
\textbf{Batch Size}  &\textbf{Epoch} & \textbf{\begin{tabular}[c]{@{}c@{}}Percentage of\\ Training Images\\  (same-different)\end{tabular}} &  \textbf{\begin{tabular}[c]{@{}c@{}}Global \\ Accuracy\end{tabular}} & \textbf{\begin{tabular}[c]{@{}c@{}}Same Room \\ Accuracy\end{tabular}} & \textbf{\begin{tabular}[c]{@{}c@{}}Different Room \\ Accuracy\end{tabular}} \\ \hline
256&7& 5\%-95\%  & 94.35\% & \textcolor{red}{96.47\%} & 92.23\% \\ \hline
256&11& 5\%-95\% & 94.24\% & 97.77\% & 90.72\% \\ \hline
256&7& 10\%-90\% & 93.04\% & 97.16\% & 88.92\% \\ \hline
256&11& 10\%-90\%  & 93.59\% & 97.96\% & 89.22\%\\ \hline
256&7& 25\%-75\%  & 92.46\% & \textcolor{green}{99.21\%} & 85.71\%\\ \hline
256&11& 25\%-75\%  & \textcolor{red}{91.78\%} & 98.81\% & \textcolor{red}{84.74\%}\\ \hline
16&7& 5\%-95\%  & \textcolor{green}{95.50\%} & 98.26\% & \textcolor{green}{92.74\%} \\ \hline
16&11& 5\%-95\% & 93.84\% & 98.83\% & 88.85\% \\ \hline
16&7& 10\%-90\% & 93.77\% & 98.13\% & 89.41\% \\ \hline
16&11& 10\%-90\%  & 94.42\% & 98.80\% & 90.05\%\\ \hline
16&7& 25\%-75\%  & 94.77\% & 99.15\% & 90.39\%\\ \hline
16&11& 25\%-75\%  & 94.08\% & 99.15\% & 89.00\%\\ \hline
\end{tabular}
\end{table}

    \begin{table}[!htb]
    \centering
\caption{Accuracy using \textbf{AlexNet} and different batch sizes.}\label{tab:AccuBSAlexNet}
\begin{tabular}{|c|c|c|c|c|c|c|c|}
\hline
\textbf{Batch Size}  &\textbf{Epoch} & \textbf{\begin{tabular}[c]{@{}c@{}}Percentage of\\ Training Images\\  (same-different)\end{tabular}} &  \textbf{\begin{tabular}[c]{@{}c@{}}Global \\ Accuracy\end{tabular}} & \textbf{\begin{tabular}[c]{@{}c@{}}Same Room \\ Accuracy\end{tabular}} & \textbf{\begin{tabular}[c]{@{}c@{}}Different Room \\ Accuracy\end{tabular}} \\ \hline
256&7& 5\%-95\%  &  \textcolor{red}{89.76\%} &\textcolor{red}{90.11\%}& \textcolor{green}{94.60\%}  \\ \hline
256&11& 5\%-95\% & 93.58\% &  94.08\%& 93.07\%  \\ \hline
256&7& 10\%-90\% & 93.77\% & 98.13\% &  89.41\%   \\ \hline
256&11& 10\%-90\%  & 94.42\% & 98.80\% & 90.05\% \\ \hline
256&7& 25\%-75\%  & 90.91\% &  97.54\%& 84.28\%  \\ \hline
256&11& 25\%-75\%   & 91.16\% &  98.92\%&  \textcolor{red}{83.39\%}  \\ \hline
16&7& 5\%-95\%  & 94.64\% & 96.25\% & 93.02\% \\ \hline
16&11& 5\%-95\% &  \textcolor{green}{95.24\%} & 98.25\% & 92.24\% \\ \hline
16&7& 10\%-90\% & 95.06\% & 98.87\% & 91.25\% \\ \hline
16&11& 10\%-90\%  & 95.07\% & 98.92\% & 91.22\%\\ \hline
16&7& 25\%-75\%  & 94.76\% & 99.10\% & 90.42\%\\ \hline
16&11& 25\%-75\%  & 94.60\% &  \textcolor{green}{99.26\%} & 89.94\%\\ \hline
\end{tabular}
\end{table}

     \vspace{0.5cm}
These tables show that relatively good performances can be achieved with some configurations. Notwithstanding that, we observe that in general terms, the same-room accuracy tends to decrease when the different-room accuracy increases and vice versa. This will be analyzed deeply in future works, but it may be due to the use of the Contrastive Loss function \citep{sun2020circle}.

\subsubsection{Influence of the architecture of the flattening layers} 
% \item[Scaled Layers Election:]\hfill \\ 

As explained in subsection \ref{subsec:Parameters}, the feature extraction layers output a matrix that is transformed to a vector in the flattening phase. Different combinations of fully connected layers are also evaluated. All these experiments have been performed training the network with a 10\% of pairs of images taken from the same room and a 90\% of pairs of images from different rooms.
 
 \begin{table}[!htb]
 \centering
\caption{Accuracy using \textbf{VGG16} and different flattening layers.}\label{tab:AccuRLVGG16}
\begin{tabular}{|c|c|c|c|c|c|}
\hline
\textbf{\begin{tabular}[c]{@{}c@{}}Flattening\\ layers\end{tabular}} & \textbf{\begin{tabular}[c]{@{}c@{}}Batch\\ Size\end{tabular}} &\textbf{Epoch} & \textbf{\begin{tabular}[c]{@{}c@{}}Global\\ Accuracy\end{tabular}} & \textbf{\begin{tabular}[c]{@{}c@{}}Same Room\\ Accuracy\end{tabular}} & \textbf{\begin{tabular}[c]{@{}c@{}}Different Room\\ Accuracy\end{tabular}} \\ \hline
500-500-5 & 16 & 7 & \textcolor{red}{93.77\%} & \textcolor{red}{98.13\%} & \textcolor{red}{89.41\%} \\ \hline
500-500-5 & 16& 11 & 94.42\% & 98.80\% & 90.05\% \\ \hline
500-500-5 & 16& 14 & 94.75\% & 99.10\% & 90.39\% \\ \hline
500-100-10 & 16& 7 & 95.76\% & 98.92\% & 92.60\% \\ \hline
500-100-10 & 16& 11 & 95.98\% & 99.11\% & 92.86\% \\ \hline
500-100-10 & 16& 14 & 95.44\% & \textcolor{green}{99.18\%} & 91.70\% \\ \hline
1000-1000-10 & 16& 7 & \textcolor{green}{96.16\%} & 98.90\% & \textcolor{green}{93.41\%} \\ \hline
1000-1000-10 & 16& 11  & 95.63\% & 99.10\% & 92.16\% \\ \hline
1000-1000-10 & 16& 14 & 95.27\% & 99.10\% & 91.44\% \\ \hline
\end{tabular}
\end{table}

 \begin{table}[!htb]
 \centering
\caption{Accuracy using \textbf{AlexNet} and different flattening layers.}\label{tab:AccuRLAlexnet}
\begin{tabular}{|c|c|c|c|c|c|}
\hline
\textbf{\begin{tabular}[c]{@{}c@{}}Flattening\\ layers\end{tabular}} & \textbf{\begin{tabular}[c]{@{}c@{}}Batch\\ Size\end{tabular}} &\textbf{Epoch} & \textbf{\begin{tabular}[c]{@{}c@{}}Global\\ Accuracy\end{tabular}} & \textbf{\begin{tabular}[c]{@{}c@{}}Same Room\\ Accuracy\end{tabular}} & \textbf{\begin{tabular}[c]{@{}c@{}}Different Room\\ Accuracy\end{tabular}} \\ \hline
500-500-5 & 16 & 7 &  \textcolor{red}{93.77\%} & \textcolor{red}{98.13\%} & 89.41\% \\ \hline
500-500-5 & 16 & 11 &  94.42\% & 98.80\% & 90.05\% \\ \hline
500-500-5 & 16 & 14 &  93.84\% & 98.68\% & \textcolor{red}{88.99\%} \\ \hline
500-100-10 & 16 & 7  & 95.31\% & 98.20\% & \textcolor{green}{92.42\%} \\ \hline
500-100-10 & 16 & 11 & 95.41\% & 98.98\% & 91.83\% \\ \hline
500-100-10 & 16 & 14 & 95.10\% & \textcolor{green}{99.06\%} & 91.15\% \\ \hline
1000-1000-10 & 16 & 7 & \textcolor{green}{95.36\%} & 98.72\% & 91.99\% \\ \hline
1000-1000-10 & 16 & 11 & 94.66\% & 98.59\% & 90.74\% \\ \hline
1000-1000-10 & 16 & 14 & 95.28\% & 99.12\% & 91.43\% \\ \hline
\end{tabular}
\end{table} 

% \vspace{0.5cm}

Tables \ref{tab:AccuRLVGG16} and \ref{tab:AccuRLAlexnet} show the results using 3 different combinations of fully connected layers. Each variation is described in Table \ref{tab:Layers_Classif}. Similar results are obtained with the 3 different variations. The best result is obtained with 3 fully connected layers with 1000-1000-10 neurons each. Finally, if we analyse jointly all the results of the room discrimination experiment, the best result is obtained using VGG16 as the feature extraction network, 3 fully connected layers (1000-1000-10), 7 epoch and a batch size of 16; with this configuration 96.16\% global accuracy is obtained: 98.90\% same room accuracy and 93.41\% different room accuracy.

 \subsection{Global Localization}
 
%   The map is composed by the set of image descriptors and their poses \(\{(f_1,\vec{p}_1),(f_2,\vec{p}_2),...,(f_N,\vec{p}_N)\}\). Each descriptor has been calculated by a Siamese Neural Network. The absolute localization is performed as a pairwise comparison between image descriptors. Given a test image $I_t$, the Siamese Neural Network outputs its corresponding descriptor $f_t$. Finally, the position of the robot is estimated by selecting the pose associated to the descriptor in the map that minimizes the distance $\|f_t-f_i\|_2$, with $i = 1,\ldots, N$.  
  
%   \textcolor{red}{SUBIR A 5.3}
 
 The global localization is performed as explained in subsection \ref{subsec:AbsoluteLoc}. The VGG16 network is employed in this task since it led to the best results in the room discrimination task. Different experiments have been performed in order to choose the best configuration. We will mainly analyze the ratio of same/different room pairs, which is the parameter that has shown the greatest influence on the results. Moreover, in this subsection we will assess the influence of the data augmentation on the results. Each pair of images is labelled according the Equation \ref{eqn:labelling_loc_global}.

\vspace{0.5cm}
First, concerning the experiment to evaluate the influence of the ratio same/different room pairs, we train the network using 8,486 pairs of images per epoch from the training dataset 1. Second, with respect to the experiment to assess the effect of the data augmentation, 977,856 pairs of images per epoch from the training dataset 2 are used. These two experiments are described in subsection \ref{subsec:influence_locglobal}. In both cases, the flattening layers are configured with 500-500-5 neurons. Moreover, subsection \ref{subsec:flattening_locglobal} evaluates the influence of the flattening layers. In this case, the training dataset 1 is used. As common parameters, we use 16 as batch size, the Stochastic Gradient Descent (SGD) optimizer, with a learning rate of 0.001 and a momentum of 0.9 and 30 epochs.

 \subsubsection{Influence of the training parameters}
 \label{subsec:influence_locglobal}

%\begin{enumerate}%[(a)]    %items (quierop ponerlos con letras)
\begin{description}

    \item[\textbf{Ratio of same/different room pairs:}] \hfill \\

Table \ref{tab:AbsLocPercentage} shows the results using VGG16 in the feature extraction part and three fully connected layers with  500-500-5 neurons in the flattening part. The training of the model has been performed with different percentages of pairs of images belonging to the same and different rooms. The results show that the lowest localization error is obtained when the training is performed using 40\% of images from the same room and 60\% of images from different rooms. Studying the results, as a general rule, training with a large percentage of image pairs from the same room deteriorates the localization error.

\begin{table}[!htb]
\centering
\caption{Average localization error (m) with \textbf{VGG16}. The table presents the global localization results with variations in the same-different ratio of training image pairs.}\label{tab:AbsLocPercentage}
\begin{tabular}{|c|c|c|c|c|c|}
\hline
\textbf{\begin{tabular}[c]{@{}c@{}}Feature \\ extraction net\end{tabular}} & \textbf{\begin{tabular}[c]{@{}c@{}}Percentage of\\ Training Images\\ (same-different)\end{tabular}} & \textbf{\begin{tabular}[c]{@{}c@{}}Global\\ Error\end{tabular}} & \textbf{\begin{tabular}[c]{@{}c@{}}Cloudy\\ Error\end{tabular}} & \textbf{\begin{tabular}[c]{@{}c@{}}Night\\ Error\end{tabular}} & \textbf{\begin{tabular}[c]{@{}c@{}}Sunny\\ Error\end{tabular}} \\ \hline
VGG16 & 80\%-20\% & \textcolor{red}{0.6284 m} & \textcolor{red}{0.1826 m} & \textcolor{red}{0.5600 m} & 0.8023 m \\ \hline
VGG16 & 70\%-30\% & 0.6035 m & 0.1753 m & 0.5375 m & 0.7706 m \\ \hline
VGG16 & 60\%-40\% & 0.6006 m & 0.1802 m & 0.4991 m & \textcolor{red}{0.8649 m} \\ \hline
VGG16 & 50\%-50\% & 0.5821 m & 0.1747 m & 0.4837 m & 0.8383 m \\ \hline
VGG16 & 40\%-60\% & \textcolor{green}{0.5097 m} & \textcolor{green}{0.1481 m} & \textcolor{green}{0.4547 m} & \textcolor{green}{0.6508 m} \\ \hline
VGG16 & 30\%-70\% & 0.5193 m & 0.1518 m & 0.4908 m & 0.6630 m \\ \hline
VGG16 & 20\%-80\% & 0.5202 m & 0.1521 m & 0.4917 m & 0.6642 m \\ \hline
\end{tabular}
\end{table}
    
   \item[\textbf{Data Augmentation:}] \hfill \\

Next, we evaluate the influence of the data augmentation on the localization task. Table \ref{tab:AbsLocPercentageDA} presents the results using the training dataset 2 (augmented) and test datasets 1, 2 and 3. For this purpose, we will start from the best configurations obtained so far and show the results according to the percentage of training image pairs.
     
% \begin{table}[H]
%     \centering
% \caption{ The table presents global localization results with variations in the same-different ratio of images and using \textbf{VGG16} and Data Augmentation. \textcolor{red}{rellenar datos que faltan, flattening layers 500-500-5}}\label{tab:AbsLocPercentageDA}
% \begin{tabular}{|c|c|c|c|c|c|c|}
% \hline
% \textbf{\begin{tabular}[c]{@{}c@{}}Feature \\ extraction net\end{tabular}} & \textbf{Images} & \textbf{\begin{tabular}[c]{@{}c@{}}Percentage Images\\ (same-different)\end{tabular}} &  \textbf{\begin{tabular}[c]{@{}c@{}}Cloudy\\ Error\end{tabular}} & \textbf{\begin{tabular}[c]{@{}c@{}}Night\\ Error\end{tabular}} & \textbf{\begin{tabular}[c]{@{}c@{}}Sunny\\ Error\end{tabular}} \\ \hline
% VGG16 & 1333440 & 50\%-50\%  & 0.0630 m & 0.4682 m & 1.1880 m \\ \hline
% VGG16 & 1333440 & 40\%-60\% & 0.0608 m & 0.4682 m & 1.2703 m \\ \hline
% VGG16 & 10667520 & 50\%-50\% & 0.0440 m & 0.3591 m & 1.2264 m \\ \hline
% VGG16 & 10667520 & 40\%-60\% & 0.0418 m & 0.3176 m & 1.0050 m \\ \hline
% VGG16 & 3666960 & 50\%-50\% & 0.0375 m & 0.3961 m & 1.8257 m \\ \hline
% VGG16 & 3666960 & 40\%-60\% & 0.0407 m & 0.4028 m & 1.3733 m \\ \hline
% VGG16 & 10756416 & 50\%-50\% &  m &  m &  m \\ \hline
% VGG16 & 10756416 & 40\%-60\% & 0.0375 m & 0.2675 m & 1.051 m \\ \hline
% VGG16 & 20534976 & 50\%-50\% &  m &  m &  m \\ \hline
% VGG16 & 20534976 & 40\%-60\% & 0.0325 m & 0.2573 m & 0.9913 m \\ \hline
% \end{tabular}
% \end{table}
 
 \begin{table}[!htb]
    \centering
\caption{Average localization error (m) with \textbf{VGG16} and data augmentation. The table presents the global localization results with variations in the same-different ratio of  training image pairs.}\label{tab:AbsLocPercentageDA}
\begin{tabular}{|c|c|c|c|c|c|c|}
\hline
\textbf{\begin{tabular}[c]{@{}c@{}}Feature \\ extraction net\end{tabular}} & \textbf{Epochs} & \textbf{\begin{tabular}[c]{@{}c@{}}Percentage of\\ Training Images\\ (same-different)\end{tabular}} &  \textbf{\begin{tabular}[c]{@{}c@{}}Cloudy\\ Error\end{tabular}} & \textbf{\begin{tabular}[c]{@{}c@{}}Night\\ Error\end{tabular}} & \textbf{\begin{tabular}[c]{@{}c@{}}Sunny\\ Error\end{tabular}} \\ \hline
VGG16 & 1 & 50\%-50\%  & \textcolor{red}{0.0630 m} & \textcolor{red}{0.4682 m} & 1.1880 m \\ \hline
VGG16 & 1 & 40\%-60\% & 0.0608 m & 0.4682 m & 1.2703 m \\ \hline
VGG16 & 4 & 50\%-50\% & 0.0375 m & 0.3961 m & \textcolor{red}{1.8257 m} \\ \hline
VGG16 & 4 & 40\%-60\% & 0.0407 m & 0.4028 m & 1.3733 m \\ \hline
VGG16 & 11 & 50\%-50\% & 0.0440 m & 0.3591 m & 1.2264 m \\ \hline
VGG16 & 11 & 40\%-60\% & 0.0418 m & 0.3176 m & 1.0050 m \\ \hline
VGG16 & 21 & 50\%-50\% & 0.0375 m & 0.2675 m & 1.051 m \\ \hline
VGG16 & 21 & 40\%-60\% & \textcolor{green}{0.0325 m} & \textcolor{green}{0.2573 m} & \textcolor{green}{0.9913 m} \\ \hline
\end{tabular}
\end{table}
 
 \end{description}
 
 \subsubsection{Influence of the architecture of the flattening layers} 
 \label{subsec:flattening_locglobal}
 To conclude the experimental section, Table \ref{tab:AbsLocScaled} shows the results after evaluating different flattening layers. In this case, a descriptor of dimension 5 works better in the global localization problem. The Siamese Neural Network is able to perform the localization with an average error of 0.5821 \textit{m} when using as the flattening layers three different fully connected layers with 500, 500 and 5 neurons.

\begin{table}[!htb]
\centering
\caption{Average localization error (m) with \textbf{VGG16} and different configurations of the flattening layers.}\label{tab:AbsLocScaled}
\begin{tabular}{|c|c|c|c|c|}
\hline
\textbf{\begin{tabular}[c]{@{}c@{}}Feature \\ extraction net \end{tabular}} & \textbf{\begin{tabular}[c]{@{}c@{}}Flattening \\ layers\end{tabular}} & \textbf{Epoch} & \textbf{\begin{tabular}[c]{@{}c@{}}Percentage of\\ Training Images\\ (same-different)\end{tabular}} & \textbf{\begin{tabular}[c]{@{}c@{}}Global\\ Error\end{tabular}} \\ \hline
VGG16& 500-500-5 & 30 & 50\%-50\% & \textcolor{green}{0.5821 m} \\ \hline
VGG16 & 1000-1000-10 & 30 & 50\%-50\% & 0.5904 m\\ \hline
VGG16 & 4096-4096-1000 & 30 & 50\%-50\% & \textcolor{red}{0.8313 m} \\ \hline
\end{tabular}
\end{table}

\subsubsection{General comparison with other methods}

Finally, the Siamese Neural Networks are compared with other previous global-appearance techniques which include the use of a single AlexNet structure and two classic analytic descriptors: HOG and gist, as described in the work by \cite{cebollada2022development}. Table \ref{tab:ComparacionMetodos} compares all the methods in a global localization task using, in all cases, the COLD-Freiburg Dataset. As we can notice, the best results in terms of localization error for cloudy and night conditions are obtained with the siamese structures proposed in the present work. The best results for the sunny illumination condition are obtained first with the single Alexnet structure and second with the siamese network without data augmentation.
     
\begin{table}[!htb]
\centering
\caption{ Comparison with other methods.}\label{tab:ComparacionMetodos}
\begin{tabular}{|l|c|c|c|}
\hline
\multicolumn{1}{|c|}{\textbf{\begin{tabular}[c]{@{}c@{}}Global-Appearance \\ Descriptor Technique\end{tabular}}} & \textbf{\begin{tabular}[c]{@{}c@{}}Cloudy\\ Error\end{tabular}} & \textbf{\begin{tabular}[c]{@{}c@{}}Night\\ Error\end{tabular}} & \textbf{\begin{tabular}[c]{@{}c@{}}Sunny\\ Error\end{tabular}} \\ \hline
Alexnet \citep{cebollada2022development}  & 0.0509 m & 0.2880 m  &  \textcolor{green}{0.3894 m}\\ 
Siamese Network (ours) & 0.1481 m  & 0.4547 m  & 0.6508 m  \\
Siamese Network + DA (ours)  & \textcolor{green}{0.0325 m} & \textcolor{green}{0.2573 m} & \textcolor{red}{0.9913 m} \\ \hline
HOG \citep{cebollada2022development} & \textcolor{red}{0.1634 m} & 0.4509 m & 0.8198 m \\
gist \citep{cebollada2022development} & 0.0519 m & \textcolor{red}{1.065 m} & 0.8839 m \\ \hline
\end{tabular}
\end{table}

\section{Conclusions}
\label{sec:conclusion}

% The objective of this paper is to address the mapping and localization tasks using omnidirectional images captured by a catadioptric vision system on board a mobile robot. For this purpose, Siamese Neural Networks are proposed due to their ability to generate a similarity function between a pair of input data.   

In this paper, a global localization method using Siamese Neural Networks has been proposed and evaluated. Localization, along with mapping, is one of the main tasks to be addressed by an autonomous mobile robot. First, an initial task of discriminating same and different rooms has been proposed in order to assess the ability of Siamese Neural Networs and know the influence of the most relevant parameters. After that, the global localization problem is addressed.

\vspace{0.5cm}

In the experiments, several well known architectures have been tested to conform the Siamese Neural Network, some of which are AlexNet, VGG11, VGG13, VGG16, VGG19, VGG11bn, VGG13bn, VGG16bn and VGG19bn. The best performance in the initial task has been achieved by VGG13 and VGG16. In general terms, the VGG architectures have provided the best results.

\vspace{0.5cm}

Apart from these feature extraction architectures, a group of Fully Connected layers have been added to carry out the conversion of the activation maps resulting from the convolutional layers to a description vector. In the present work, different sizes of the Fully Connected layers have been studied, as well as the size of the final descriptor. For the initial task, the performance of the network is slightly higher when the Fully Connected layers sizes are 1000-1000-10. In contrast, in  the global localization, the localization error decreases drastically in those networks that have a set of Fully Connected layers of size 500-500-5 neurons.

\vspace{0.5cm}

The training parameter that contributes most to the performance of the network is the percentage of image pairs belonging to the same and different rooms. In this sense, there is a correlation between the percentage of images of same/different room and its respective accuracy, i.e., when the percentage of pairs of images in the same room increases, its associated accuracy also does and a similar effect occurs with the different room category. Furthermore, when the same room accuracy increases, the different room accuracy decreases, and vice versa. This situation may be caused by the Contrastive Loss function which has an associated lack of flexibility in the optimization. Other loss functions used in other applications could improve localization results, such as Circle Loss \citep{circle_loss} and will be considered in future studies. 

% Furthermore, this initial task has the capability of Siamese Neural Networks to approach the global localization problem.
\vspace{0.5cm}

In addition, a data augmentation technique has been proposed in order to improve the performance of the network. The proposed effects try to simulate real operation conditions. In addition, a set of effects specially designed to increase the robustness against changes of the lighting conditions in the scene have been generated. As for the results obtained, the performance of the network is especially benefited when working in cloudy and night lighting conditions. In the case of the cloudy lighting condition, when the training is performed with data augmentation, the average localization error is reduced around 12 centimeters. As for the night illumination condition, the average error is reduced around 20 centimeters. On the contrary, in sunny illumination condition the average localization error increases 34 centimeters when data augmentation is used.

\vspace{0.5cm}

As future works, the proposed localization techniques will be extended to outdoor environments, which are more challenging because of their unstructured and changing conditions. In addition, other types of sensors will be considered to carry out the localization robustly, such as LiDAR.

%% Meter de alguna forma

% Other loss functions used in other applications could have improved localization results, such as Circle Loss \citep{circle_loss}. 

\section*{Acknowledgements}

The Ministry of Science, Innovation and Universities (Spain) has supported this work through “Ayudas para la Formación de Profesorado Universitario” (FPU21/04969). This work is also part of the project TED2021-130901B-I00, funded by MCIN/AEI/10.13039/501100011033 and by the European Union “NextGenerationEU”/PRTR, and of the project PID2020-116418RB-I00 funded by MCIN/AEI/10.13039/501100011033.
%\noindent This work has been supported by the Generalitat Valenciana and the FSE through the grant ACIF/2018/224, by the Spanish Government through the project DPI 2016-78361-R (AEI/FEDER, UE): “Creación de mapas mediante métodos de apariencia visual para la navegación de robots” and by Generalitat Valenciana through the project AICO/2019/031: “Creación de modelos jerárquicos y localización robusta de robots móviles en entornos sociales”.



\bibliography{sn-bibliography}% common bib file


\end{document}